\pdfoutput=1
\documentclass[11pt]{article}

\usepackage{emnlp2021}

\usepackage{times}
\usepackage{latexsym}

\usepackage{amssymb}
\usepackage{amsmath}
\usepackage[capitalise]{cleveref}

\usepackage{booktabs}
\usepackage{multirow}
\usepackage{pgfplots}
\usepackage{longfbox}

\usepackage[T1]{fontenc}

\usepackage[utf8]{inputenc}

\usepackage{microtype}

\usepackage{todonotes}

\usepackage{tikz}
\usepackage{collcell}
\usepackage{adjustbox}
\usepackage{subcaption}

\newcommand{\orthoprobe}[0]{\emph{Orthogonal Structural Probe}}

\newcommand{\scalingvec}[0]{\emph{Scaling Vector}}

\newcommand{\orthotransf}[0]{\emph{Orthogonal Transformation}}

\usepackage[normalem]{ulem}

\def\daviddel#1{\bgroup\markoverwith{\textcolor{green}{\rule[0.4ex]{2pt}{3pt}}}\ULon{#1}}

\def\ZKdel#1{\bgroup\markoverwith{\textcolor{green!60!black!100}{\rule[0.4ex]{2pt}{3pt}}}\ULon{#1}}

\newcommand*{\MinNumber}{-.3}%
\newcommand*{\MidNumber}{0.0} %
\newcommand*{\MaxNumber}{.05}%

\newcommand{\ApplyGradient}[1]{%
        \ifdim #1 pt > \MidNumber pt
            \pgfmathsetmacro{\PercentColor}{max(min(100.0*(#1 - \MidNumber)/(\MaxNumber-\MidNumber),100.0),0.00)} %
        
            \noindent\colorbox{green!\PercentColor!white}{#1}
        \else
            \pgfmathsetmacro{\PercentColor}{max(min(100.0*(\MidNumber - #1)/(\MidNumber-\MinNumber),100.0),0.00)} 
            \noindent\colorbox{red!\PercentColor!white}{#1}
        \fi
}

\title{Examining Cross-lingual Contextual Embeddings with Orthogonal Structural Probes}

\author{Tomasz Limisiewicz \and David Mare\v{c}ek \\
    Institute of Formal and Applied Linguistics, Faculty of Mathematics and Physics \\
    Charles University, Prague, Czech Republic \\
  \texttt{\{limisiewicz, marecek\}@ufal.mff.cuni.cz}
}
\begin{document}
\maketitle

\begin{abstract}

State-of-the-art contextual embeddings are obtained from large language models available only for a few languages. For others, we need to learn representations using a multilingual model. There is an ongoing debate on whether multilingual embeddings can be aligned in a space shared across many languages.
The novel \orthoprobe~\cite{limisiewicz-2020-introducing} allows us to answer this question for specific linguistic features and learn a projection based only on mono-lingual annotated datasets.
We evaluate syntactic (UD) and lexical (WordNet) structural information encoded in  \textsc{mBERT}'s contextual representations for nine diverse languages.\footnote{English, Spanish, Slovene, Indonesian, Chinese, Finnish, Arabic, French, and Basque}
We observe that for languages closely related to English, no transformation is needed. The evaluated information is encoded in a shared cross-lingual embedding space. For other languages, it is beneficial to apply orthogonal transformation learned separately for each language.
We successfully apply our findings to zero-shot and few-shot cross-lingual parsing.

\end{abstract}

\section{Introduction}
The representation learned by language models has been successfully applied in various NLP tasks. Multilingual pre-training allows utilizing the representation for various languages, including low-resource ones. There is an open discussion about to what extent contextual embeddings are similar across languages \cite{sogaard-etal-2018-limitations, hartmann-2019-comparing, vulic-2020-good}. The motivation for our work is to answer: \textbf{Q1} Is linguistic information uniformly encoded in the representations of various languages? And if this assumption does not hold: \textbf{Q2} Is it possible to learn orthogonal transformation to align the embeddings?

We probe for the syntactic and lexical structures encoded in multilingual embeddings with the new \orthoprobe\emph{s}~\citep{limisiewicz-2020-introducing}. Previously, \citet{chi-etal-2020-finding} employed \emph{structural probing} \citep{hewitt-manning-2019-structural} to evaluate cross-lingual syntactic information in \textsc{mBERT} and visualize how it is distributed across languages. Our approach's advantage is learning an orthogonal transformation that maps the embeddings across languages based on mono-lingual linguistic information: dependency syntax and lexical hypernymy. This new capability allows us to test different probing scenarios. We measure how adding assumptions of isomorphism and uniformity of the representations across languages affect probing results to answer our research questions.  






\section{Related Work}
\paragraph{Probing} 
It is a method of evaluating linguistic information encoded in pre-trained NLP models. Usually, a simple classifier for the probing task is trained on the frozen model's representation \citep{linzen2016assessing, belinkov2017neural, blevins2018deep}.  The work of \citet{hewitt-manning-2019-structural} introduced structural probes that linearly transform contextual embeddings to approximate the topology of dependency trees.
\citet{limisiewicz-2020-introducing} proposed new structural tasks and introduced orthogonal constraint allowing to decompose projected embeddings into parts correlated with specific linguistic features.
\citet{kulmizev-etal-2020-neural} probed different languages to examine what type of syntactic dependency annotation is captured in an LM.
\citet{hall-maudslay-etal-2020-tale} modify the loss function, improving syntactic probes' ability to parse.

\paragraph{Cross-lingual embeddings}
There is an essential branch of research studying relationships of embeddings across languages. \citet{mikolov-2013-exploiting} showed that distributions of the word vectors in different languages could be aligned in shared space. Following research analyzed various methods of aligning cross-lingual static embeddings \citep{faruqui-dyer-2014-improving, artetxe-etal-2016-learning, smith-etal-2017-offline} and gradually dropped the requirement of parallel data for alignment \citep{artetxe-etal-2018-robust, zhang-etal-2017-earth, lample-etal-2018-word}. 


 Significant attention was also devoted to the analysis of multilingual and contextual embeddings of \textsc{mBERT} \cite{pires-etal-2019-multilingual, libovicky-etal-2020-language}. There is also no conclusive answer to whether the alignment of such representations is beneficial to cross-lingual transfer. \citet{wang-etal-2019-cross} show that the alignment facilitates zero-shot parsing, while results of \citet{wu-dredze-2020-explicit} for multiple tasks put in doubt the benefits of the alignment.

\section{Method}

The \emph{Structural Probe} \citep{hewitt-manning-2019-structural}  is a gradient optimized linear projection of the contextual word representations produced by a pre-trained neural model (e.g. \textsc{BERT} \citet{devlin-etal-2019-bert}, \textsc{ELMo} \citet{peters2018deep}).

In a \emph{Distance Probe}, the Euclidean distance between projected word vectors approximates the distance between words in a dependency tree:
\begin{equation}
    \label{eqn:distance-probe}
    d_B(h_i,h_j)^2 = (B(h_i - h_j))^T(B(h_i - h_j)),
\end{equation}
$B$ is the \emph{Linear Transformation} matrix and $h_i$, $h_j$ are the vector representations of words at positions $i$ and $j$.

Another type of a probe is a \emph{Depth Probe}, where the token's depth in a dependency tree is approximated by the Euclidean norm of a projected word vector:
\begin{equation}
    \label{eqn:depth-probe}
    ||h_i||_B^2 = (Bh_i)^T(Bh_i)
\end{equation}

\paragraph{Orthogonal Structural Probes}

\citet{limisiewicz-2020-introducing} proposed decomposing matrix $B$ and then gradient optimizing a vector and orthogonal matrix.
The new formulation of an \emph{Orthogonal Distance Probe} is\footnote{Reformulation of an \emph{Orthogonal Depth Probe} is analogical.}:
\begin{equation}
\begin{split}
    &d_{\bar{d}V^T}(h_i,h_j)^2 \\
    &=(\bar{d} \odot V^T  (h_i - h_j))^T( \bar{d} \odot V^T (h_i - h_j)),
\end{split}
\end{equation}

where $V$ is an orthogonal matrix (\orthotransf) and $\bar{d}$ is a \scalingvec, which can be changed during optimization for each task to allow multi-task joint probing. 

This procedure allowed optimizing a separate \scalingvec~$\bar{d}$  for a specific objective, allowing probing for multiple linguistic tasks simultaneously. In this work, an individual \orthotransf~$V$ is trained for each language, facilitating multi-language probing.
This approach assumes that the representations are isomorphic across languages; we examine this claim in our experiments. 




Our implementation is available at GitHub: \url{https://github.com/Tom556/OrthogonalTransformerProbing}.

\section{Experiments}

\begin{table*}[!ht]
\setlength{\tabcolsep}{2 pt}
\fboxset{rounded,
        border-color=gray,
        padding={0.0 pt,0.0 pt}}
\centering
\small
\begin{tabular}{@{}l|ccccccccc|ccc@{}}
\toprule
\textbf{Approach} & \textbf{EN} & \textbf{ES} & \textbf{SL} & \textbf{ID} & \textbf{ZH} & \textbf{FI} & \textbf{AR} & \textbf{FR} & \textbf{EU} & \multicolumn{3}{c}{\textbf{AVERAGE}} \\
 &  & &  & &  &  &  &  &  & \textbf{I-E} & \textbf{N-I-E} & \textbf{All} \\\midrule
 \multicolumn{13}{c}{\textbf{Dependency Distance Spearman's Correlation}} \\ \midrule
\textsc{In-Lang} & .812 & .858 & .857 & .841 & .830 & .788 & .838 & .856 & .769 & .846 & .813 & .828 \\
\textcolor{blue}{Chi et al.} & .817 & .859 & - & .807 &  .777 & .812 & .822 & .864 & - & .847 & .805 & .823 \\ \midrule
$\Delta$ \textsc{MappedL} & \ApplyGradient{.000} & \lfbox[tight]{\ApplyGradient{-.001}} & \ApplyGradient{.001} & \ApplyGradient{-.003} & \ApplyGradient{.000} & \ApplyGradient{.001} & \ApplyGradient{-.001} & \lfbox[tight]{\ApplyGradient{-.002}} & \ApplyGradient{.001} & \ApplyGradient{-.001} & \ApplyGradient{.000} & \ApplyGradient{.000} \\ \midrule
$\Delta$ \textsc{AllL} & \ApplyGradient{.000} & \lfbox[tight]{\ApplyGradient{-.007}} & \lfbox[tight]{\ApplyGradient{-.006}} & \lfbox[tight]{\ApplyGradient{-.013}} & \lfbox[tight]{\ApplyGradient{-.039}} & \ApplyGradient{.000} & \lfbox[tight]{\ApplyGradient{-.027}} & \lfbox[tight]{\ApplyGradient{-.006}} & \lfbox[tight]{\ApplyGradient{-.032}} & \lfbox[tight]{\ApplyGradient{-.005}} & \lfbox[tight]{\ApplyGradient{-.022}} & \lfbox[tight]{\ApplyGradient{-.015}}\\ 
\textcolor{blue}{Chi et al.} & \ApplyGradient{-.011} & \ApplyGradient{-.011} & - & \ApplyGradient{-.018} & \ApplyGradient{-.060} & \ApplyGradient{-.010} & \ApplyGradient{-.037} & \ApplyGradient{-.011} & - & \ApplyGradient{-.011} & \ApplyGradient{-.031} & \ApplyGradient{-.023} \\ \midrule
\multicolumn{13}{c}{\textbf{Dependency Depth Spearman's Correlation}} \\ \midrule
\textsc{In-Lang} & .843 & .868 & .867 & .855 & .844 & .822 & .865 & .877 & .797 & .864 & .837 & .849 \\ 
$\Delta$ \textsc{MappedL} & \lfbox[tight]{\ApplyGradient{-.004}} & \lfbox[tight]{\ApplyGradient{-.003}} & \lfbox[tight]{\ApplyGradient{-.002}} & \ApplyGradient{-.002} & \ApplyGradient{.000} & \lfbox[tight]{\ApplyGradient{-.002}} & \ApplyGradient{.001} & \ApplyGradient{-.002} & \ApplyGradient{-.001} & \lfbox[tight]{\ApplyGradient{-.002}} & \lfbox[tight]{\ApplyGradient{-.001}} & \lfbox[tight]{\ApplyGradient{-.002}} \\ 
$\Delta$ \textsc{AllL} & \lfbox[tight]{\ApplyGradient{-.006}} & \lfbox[tight]{\ApplyGradient{-.007}} & \lfbox[tight]{\ApplyGradient{-.008}} & \lfbox[tight]{\ApplyGradient{-.011}} & \lfbox[tight]{\ApplyGradient{-.035}} & \lfbox[tight]{\ApplyGradient{-.005}} & \lfbox[tight]{\ApplyGradient{-.031}} & \lfbox[tight]{\ApplyGradient{-.010}} & \lfbox[tight]{\ApplyGradient{-.031}} & \lfbox[tight]{\ApplyGradient{-.008}} & \lfbox[tight]{\ApplyGradient{-.023}} & \lfbox[tight]{\ApplyGradient{-.016}} \\ \midrule
\multicolumn{13}{c}{\textbf{Lexical Distance Spearman's Correlation}} \\ \midrule
\textsc{In-Lang} & .756 & .841 & .639 & .719 & .800 & .657 & .733 & .794 & .679 & .757 & .717 & .735 \\ 
$\Delta$ \textsc{MappedL} & \ApplyGradient{-.003} & \ApplyGradient{.005} & \ApplyGradient{-.011} & \ApplyGradient{-.001} & \lfbox[tight]{\ApplyGradient{.010}} & \ApplyGradient{.001} & \lfbox[tight]{\ApplyGradient{.042}} & \ApplyGradient{.001} & \ApplyGradient{-.008} & \ApplyGradient{-.002} & \lfbox[tight]{\ApplyGradient{.009}} & \ApplyGradient{.004}\\ 
$\Delta$ \textsc{AllL} & \lfbox[tight]{\ApplyGradient{-.038}} & \lfbox[tight]{\ApplyGradient{-.025}} & \lfbox[tight]{\ApplyGradient{-.042}} & \lfbox[tight]{\ApplyGradient{-.051}} & \lfbox[tight]{\ApplyGradient{-.014}} & \lfbox[tight]{\ApplyGradient{-.043}} & \ApplyGradient{.025} & \lfbox[tight]{\ApplyGradient{-.013}} & \lfbox[tight]{\ApplyGradient{-.063}} & \lfbox[tight]{\ApplyGradient{-.030}} & \lfbox[tight]{\ApplyGradient{-.029}} & \lfbox[tight]{\ApplyGradient{-.030}} \\ \midrule
\multicolumn{13}{c}{\textbf{Lexical Depth Spearman's Correlation}} \\ \midrule
\textsc{In-Lang} & .853 & .881 & .779 & .852 & .875 & .784 & .906 & .844 & .842 & .839 & .850 & .845 \\ 
$\Delta$ \textsc{MappedL} & \ApplyGradient{.004} & \ApplyGradient{-.005} & \lfbox[tight]{\ApplyGradient{.013}} & \lfbox[tight]{\ApplyGradient{-.011}} & \ApplyGradient{.006} & \lfbox[tight]{\ApplyGradient{.023}} & \ApplyGradient{-.024} & \ApplyGradient{.007} & \lfbox[tight]{\ApplyGradient{.021}} & \lfbox[tight]{\ApplyGradient{.004}} & \ApplyGradient{.005} & \ApplyGradient{.005} \\ 
$\Delta$ \textsc{AllL} & \lfbox[tight]{\ApplyGradient{-.027}} & \lfbox[tight]{\ApplyGradient{-.048}} & \lfbox[tight]{\ApplyGradient{-.040}} & \lfbox[tight]{\ApplyGradient{-.124}} & \lfbox[tight]{\ApplyGradient{-.068}} & \ApplyGradient{-.006} & \lfbox[tight]{\ApplyGradient{-.305}} & \lfbox[tight]{\ApplyGradient{-.032}} & \lfbox[tight]{\ApplyGradient{-.020}} & \lfbox[tight]{\ApplyGradient{-.037}} & \lfbox[tight]{\ApplyGradient{-.103}} & \lfbox[tight]{\ApplyGradient{-.079}}\\ \midrule
\end{tabular}
\caption{Spearman's correlation between gold and predicted depths and distances.  $\Delta$  denotes the differences from \textsc{In-Lang} results. Each of our results is an average of 6 randomly initialized probing experiments. \lfbox[tight]{Statistically significant} differences are circled.
The three last columns present averages for \textbf{I}ndo-\textbf{E}uropean, \textbf{N}on-\textbf{I}ndo-\textbf{E}uropean, and all languages. The evaluation is not zero-shot, we use data in a target language. Correlations for dependency distance are compared with \emph{Standard Structural Probes} reported by \citet{chi-etal-2020-finding}. }
\label{tab:spearman-results}
\end{table*}

We examine vector representations obtained from multilingual cased \textsc{BERT} \cite{devlin-etal-2019-bert}. 

\subsection{Data and Probing Objectives}

We probe for syntactic structure annotated in Universal Dependencies treebanks \cite{nivre-etal-2020-ud} and for lexical hypernymy trees from WordNet \cite{miller-1995-wordnet}.
We optimize depth and dependency probes in both types of structures jointly.

For both dependency and lexical probes, we use sentences from UD treebanks in nine languages. For each treebank, we sampled 4000 sentences to diminish the effect of varying size datasets in probe optimization. Lexical depths and distances for each sentence are obtained from hypernymy trees that are available for each language in Open Multilingual Wordnet \cite{bond-foster-2013-linking}.\footnote{List of all the datasets used in this work can be found in Appendix.}

\paragraph{Choice of Layers}
We probe the representations of the 7th layer for dependency information and representations of the 5th layer for lexical information. These layers achieve the highest performance for the respective features. 

\subsection{Multilingual Evaluation}

We utilize the new joint optimization capability of \orthoprobe\emph{s} to analyze how the encoding of linguistic phenomena are expressed across different languages in \textsc{mBERT} representations.

To answer our research question, we evaluate three settings of multilingual \orthoprobe~training. The approaches are sorted by expressiveness; the most expressive one makes the weakest assumption about the likeness of representations across languages:

\paragraph{\textsc{In-Lang} no assumption}We train a separate instance of \orthoprobe~for each language. Neither \scalingvec~nor \orthotransf~is shared between languages.

\paragraph{\textsc{MappedLangs} isomorphity assumption} We train a shared \scalingvec~for each probing task and a separate \orthotransf~per language. If the embedding subspaces are orthogonal across languages, the orthogonal mapping will be learned during probe training, and the setting will achieve similar results as the previous one.

\paragraph{\textsc{AllLangs}: uniformity assumption} Both the \scalingvec~and \orthotransf~are shared across languages. If the same embedding subspace encodes the probed information across languages, the results of this setting will be on par with the first approach.


The first and the last approach was proposed analyzed for \emph{Structural Probes} by \citet{chi-etal-2020-finding}. \textsc{MappedLangs} setting is possible thanks to the new probing formulation of \citet{limisiewicz-2020-introducing}. For evaluation, we compute Spearman's correlations between predicted and gold depths and distances. 
In this evaluation, we use supervision for a target language.
Furthermore, we analyze the impact of two language-specific features on the results: a) size of the \textsc{mBERT} training corpus in a given language; b) typological similarity to English. The former is expressed in the number of tokens in Wikipedia. The latter is a Hamming similarity between features in WALS \citep{dryer-haspelmath-2013-wals}.\footnote{In this work, we consider all the features in the areas: Nominal Categories, Verb Categories, and Lexicon for computing a lexical typological similarity, and features in the areas: Nominal Syntax, Word Order, Simple Clauses, and Complex Sentences as a syntactic typological similarity. Each area contains multiple typological features.}

\subsection{Zero- and Few-shot Parsing}

We extract directed trees from the predictions of dependency probes. For that purpose, we employ the Maximum Spanning Tree algorithm on the predicted distances and the algorithm's extension of \citet{kulmizev-etal-2020-neural} to extract directed trees based on predicted depths.

We examine cross-lingual transfer for parsing sentences in Chinese, Basque, Slovene, Finnish, and Arabic. For each of them, we train the probe on the remaining eight languages. In a few-shot setting, we also optimize on 10 to 1000 examples from the target language. 


\section{Results}
\label{sec:results}
\begin{figure}[t]
    \centering
    \resizebox{\linewidth}{!}{\input{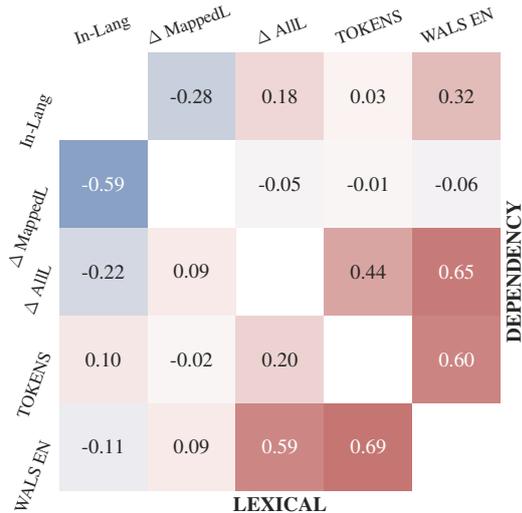}}
    \caption{Pearson's correlation between results from \cref{tab:spearman-results} for each language and two language-specific features: typological similarity to English and number of tokens in Wikipedia. Correlations for dependency probes are in the upper-right triangle and for lexical probes in the lower-left triangle.}
    \label{fig:pearson-features}
\end{figure}

\begin{table}[t]
\setlength{\tabcolsep}{5 pt}
\centering
\small
\begin{tabular}{@{}l|c|ccccc@{}}
\toprule
                        & N                     & ZH    & EU    & SL    & FI    & AR    \\ \midrule
\textcolor{blue}{Lauscher+}*               &  \parbox[t]{1mm}{\multirow{6}{*}{\rotatebox[origin=c]{90}{0}}}    & 51.41 & 50.31 & -     & 65.66 & 44.46 \\
\textcolor{blue}{Wang et al.}             &                       & -     & -     & 67.86 & 65.45 & -     \\
+CLBT**                   &                       & -     & -     & 69.04 & 67.96 & -     \\
+FT* **                     &                       & -     & -     & 69.16 & \textbf{69.16} & -     \\
\textsc{MappedL}         &                       & 34.44 & 39.10 & 35.44 & 37.33 & 40.95 \\
\textsc{AllL}         &                       & \textbf{52.92} & \textbf{58.77} & \textbf{70.76} & 64.60 & \textbf{57.47} \\ \midrule
\textcolor{blue}{Lauscher+}*               & \parbox[t]{1mm}{\multirow{3}{*}{\rotatebox[origin=c]{90}{10}}}    & 57.73 & 57.23 & -     & 65.13 & 71.00 \\
\textsc{MappedL}         &                       & 37.01 & 39.63 & 35.77 & 40.15 & 36.81 \\
\textsc{AllL}         &                       & 53.12 & 58.51 & 70.85 & 64.98 & 68.59 \\ \midrule
\textcolor{blue}{Lauscher+}*               & \parbox[t]{1mm}{\multirow{3}{*}{\rotatebox[origin=c]{90}{50}}}    & 66.78 & 66.73 & -     & 69.26 & 75.84  \\
\textsc{MappedL}         &                       & 45.07 & 50.02 & 55.09 & 49.32 & 57.77 \\
\textsc{AllLangs}         &                       & 53.63 & 59.07 & 70.43 & 65.02 & 68.81 \\ \midrule
\textcolor{blue}{Lauscher+}*               & \parbox[t]{1mm}{\multirow{3}{*}{\rotatebox[origin=c]{90}{100}}}   & 69.91 & 65.70 & -     & 70.25 & 78.50  \\
\textsc{MappedL}         &                       & 50.27 & 56.07 & 60.00 & 52.86 & 62.36 \\
\textsc{AllL}         &                       & 53.71 & 60.23 & 70.54 & 64.83 & 68.71 \\ \midrule
\textcolor{blue}{Lauscher+}*               & \parbox[t]{1mm}{\multirow{3}{*}{\rotatebox[origin=c]{90}{1000}}} & 80.12 & 74.75 & -     & 78.00 & 83.85 \\
\textsc{MappedL}         &                       & 60.57 & 65.98 & 72.81 & 63.80 & 68.85 \\
\textsc{AllL}         &                       & 57.17 & 63.49 & 72.35 & 66.05 & 69.57 \\ \bottomrule
\end{tabular}
\caption{UAS of extracted dependency trees. Our two approaches are compared to the previous works that use a biaffine parser \cite{lauscher-etal-2020-zero, wang-etal-2019-cross}. We probed the representations of the 7th layer. *): fine-tuning of \textsc{mBERT} is used. **): the multilingual dictionary is used to align the embeddings.}
\label{tab:zs-fs-uas}
\end{table}



\paragraph{Sperman's correlation}

Using \textsc{In-Lang} probes for each language gives high Spearman's correlations across the languages. The \textsc{MappedLangs} approach brings only a slight difference for most of the configuration while imposing uniformity constraint (\textsc{AllLangs}) deteriorates the results for some of the languages, as shown in \cref{tab:spearman-results}. The drop in correlation is especially high for Non-Indo-European languages (except for lexical distance where the difference between Indo-European and Non-Indo-European groups is small).

In \cref{fig:pearson-features}, we present the Pearson's correlations between results from \cref{tab:spearman-results} and two language-specific features. 
The key observation is that topological similarity to English is strongly correlated with $\Delta$\textsc{AllLangs}. Hence, a shared probe achieves relatively good for English, Spanish, and French.  It shows that lexical and dependency information is uniformly distributed in the embedding space for those languages. We bear in mind that the European languages are over-represented in the \textsc{mBERT}'s pre-training corpus. However, the size of pre-training corpora is correlated to a lesser extent with $\Delta$\textsc{AllLangs} than WALS similarity, suggesting that the latter has a more prominent role than the former. There is no significant correlation between $\Delta$\textsc{MappedLangs} and typological similarity; the embeddings of diverse languages can be similarly well mapped into a shared space. Notably, we observe that some languages with the lower performance of \textsc{In-Lang} probes can benefit from mapping (e.g., Slovene, Finnish, and Basque in the lexical depth). We view it as a benefit of cross-lingual transfer from more resourceful languages.

\paragraph{Zero-shot Parsing}

 For all languages except Finnish in zero-shot configuration, our \textsc{AllLangs} approach is better than other works that utilize a biaffine parser \cite{dozat-manning-2016-biaffine} on top of \textsc{mBERT} representations, shown in \cref{tab:zs-fs-uas}. Without any supervision, our \textsc{MappedLangs} approach performs poorly because mapping cannot be learned effectively. When some annotated data is added to the training, the difference between \textsc{AllLangs} and \textsc{MappedLanngs} decreases. We observe that between 100 and 1000 training samples are needed to learn the \orthotransf~effectively. Also, with higher supervision, we observe that the results reported by \cite{lauscher-etal-2020-zero} notably outperform our approach. The outcome was anticipated because they fine-tune \textsc{mBERT} and use biaffine with a larger capacity than a probe. For their approach, the introduction of even small supervision is more advantageous than for probing.

\section{Conclusions}

We propose an effective way to multilingually probe for syntactic dependency (UD) and lexical hypernymy (WordNet). Our algorithm learns probes for multiple tasks and multiple languages jointly. The formulation of \orthoprobe~allows learning cross-lingual transformation based on mono-lingual supervision. Our comparative evaluation indicates that the evaluated information is similarly distributed in the \textsc{mBERT}'s representations for languages typologically similar to English: Spanish, French, and Finnish. We show that aligning the embeddings with \orthotransf~improves the results for other examined languages, suggesting that the representations are isomorphic. We show that the probe can be utilized in zero- and few-shot parsing. The method achieves better UAS results for Chinese, Slovene, Basque, and Arabic in a zero-shot setting than previous approaches, which use a more complex biaffine parser.

\paragraph{Limitations}
In our choice of languages, we wanted to ensure diversity. Nevertheless, four of the analyzed languages belong to an Indo-European family that could facilitate finding shared encoding subspace for those languages.

\section*{Acknowledgments}
We thank anonymous EMNLP reviewers for their valuable comments and suggestions for improvement. 
This work has been supported by grant 338521 of the Charles University Grant Agency
and by Progress Q48 grant of Charles University.
We have been using language resources and tools developed, stored, and distributed by the LINDAT/CLARIAH-CZ project of the Ministry of Education, Youth and Sports of the Czech Republic (project LM2018101). 

\bibliography{anthology,custom}
\bibliographystyle{acl_natbib}

\newpage
\appendix

\section{WALS similarities}

\begin{figure}[h]
    \centering
    \resizebox{1.0\linewidth}{!}{\input{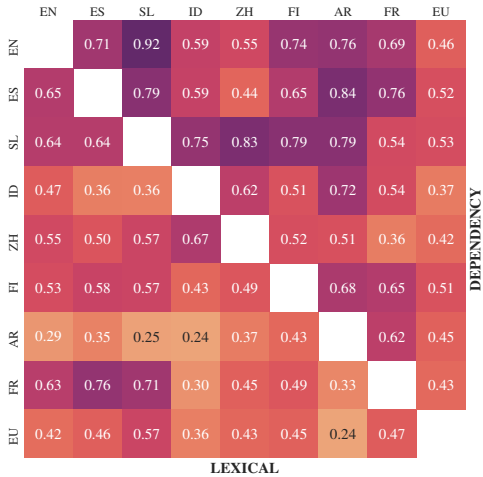}}
    \caption{Typological (WALS) similarities between languages. Dependency similarities in the upper-right triangle and lexical similarities in the lower-left triangle.}
    \label{fig:wals-sim}
\end{figure}

\begin{figure*}[!h]
    \centering
    \begin{subfigure}{0.8\linewidth}
        \resizebox{1.0\linewidth}{!}{\input{figures/dep_sim_features.pgf}}
        \caption{Dependency}
    \end{subfigure}
    \begin{subfigure}{0.8\linewidth}
        \resizebox{1.0\linewidth}{!}{\input{figures/lex_sim_features.pgf}}
        \caption{Lexical}
    \end{subfigure}
    \caption{Pearson's correlation between WALS similarity to a specific language and $\Delta$\textsc{AllLangs}, the number of tokens in Wikipedia. ``IE avg.'' stands for average similarity to analyzed Indo-European languages, i.e., English, Spanish, French, Slovene.}
    \label{fig:wals-vs-tokens-alll}
\end{figure*}

In \cref{fig:wals-sim}, we present typological similarities between languages. Bases on \cref{fig:wals-vs-tokens-alll} we observe that typological similarity to languages related to English: Spanish, Finnish, French is correlated to $\Delta$\textsc{AllLangs}. Moreover, the correlation between similarity to these languages and the number of tokens in Wikipedia is smaller than for English\footnote{English is especially over-represented in the pre-trained corpus}. It supports our claim that typological similarity is more important for uniformity assumption than the size of the pre-training corpus.

\section{Pre-training corpus size}

Sizes of Wikipedia in eight analyzed languages are presented in \cref{tab:wiki-size}.

\begin{table}[h]
\centering
\begin{tabular}{@{}c|rr@{}}
\toprule
Language & \multicolumn{1}{c}{Articles} & \multicolumn{1}{c}{Tokens} \\ \midrule
English    & 6,171,405 & 2,622,505,044 \\
French     & 2,255,469 & 823,362,731   \\
Spanish    & 1,631,829 & 688,970,215   \\
Chinese    & 1,151,113 & 269,492,468   \\
Arabic     & 1,069,379 & 169,126,089   \\
Finnish    & 494,487   & 98,712,322    \\
Indonesian & 547,825   & 96,356,452    \\
Basque     & 365,301   & 46,487,007    \\
Slovene    & 169,604   & 42,511,205    \\ \bottomrule
\end{tabular}
\caption{The number of articles and tokens in Wikipedia for analyzed languages. The data come from \url{https://github.com/mayhewsw/multilingual-data-stats/tree/main/wiki}}
\label{tab:wiki-size}
\end{table}

\section{Datasets}
In \cref{tab:datasets} we aggregate all the datasets used in our experiments.

\begin{table*}[h]
\centering
\small
\begin{tabular}{@{}l|ll|ll@{}}
\toprule
& \multicolumn{2}{c|}{Dependency} & \multicolumn{2}{c}{Lexical} \\ 
Language & Name & Reference & Name & Reference \\ \midrule
English    & EWT & \citet{silveira-2014-gold} & Princeton Wordnet & \citet{miller-1995-wordnet} \\
French     & GSD & \citet{mcdonald-etal-2013-universal} & Wordnet Libre du Français & \citet{Sagot:Fiser:2008}  \\
Spanish    & Ancora & \citet{TauleMR08} & Multilingual Central Repository &  \citet{Gonzalez-Agirre:Laparra:Rigau:2012} \\
Chinese    & GSD & \citet{mcdonald-etal-2013-universal} & Chinese Open Wordnet & \citet{Wang:Bond:2013} \\
Arabic     & PADT & \citet{Zemnek2008DependencyT} & Arabic WordNet &  \citet{Black:Elkateb:Rodriguez:Alkhalifa:Vossen:Pease:Bertran:Fellbaum:2006}\\
Finnish    & TDT   & \citet{HaverinenNVLKMOSG14} & FinnWordNet & \citet{Linden:Carlson:2010} \\
Indonesian & GSD   & \citet{mcdonald-etal-2013-universal} & Wordnet Bahasa & \citet{Noor:Sapuan:Bond:2011} \\
Basque     & BDT   & \citet{aranzabe-2015-bdt} & Multilingual Central Repository &  \citet{Pociello:Agirre:Aldezabal:2011} \\
Slovene    & SSJ   & \citet{dobrovoljc-etal-2017-universal} & sloWNet & \citet{Fiser:Novak:Eejavec:2012} \\ \bottomrule
\end{tabular}
\caption{The datasets used for training dependency and lexical probes.}
\label{tab:datasets}
\end{table*}

\section{Information separation}

In line with the findings of \citet{limisiewicz-2020-introducing} we have observed that in multilingual setting \orthoprobe \emph{s} disentangle the subspaces responsible for encoding lexical and dependency structures \cref{tab:l-6-separation}.

\begin{table}[!h]
\centering
\small
    \begin{tabular}{ll|cc|cc}
    & & \multicolumn{2}{c}{DEP} & \multicolumn{2}{|c}{LEX}  \\
    & & \rotatebox{90}{Depth} & \rotatebox{90}{Dist.} & \rotatebox{90}{Depth} & \rotatebox{90}{Dist.} \\\hline
    \multirow{5}{*}{\rotatebox{90}{DEP}} &  &  &   &   &   \\
    & Depth & 98 & 65  & 1  & 0     \\
     & &  &   &   &   \\
    & Dist. &  & 142 & 0  & 0     \\ 
     &  &  &   &   &   \\ 
    \hline
    \multirow{5}{*}{\rotatebox{90}{LEX}} &  &  &   &   &  \\
    & Depth &  &   & 22 & 13     \\
    & &  &   &   &  \\
    & Dist. &  &  & & 58  \\ 
    & &  &   &   &  \\ \hline

    \end{tabular}
    \caption{The number of shared dimensions selected by \scalingvec~after the joint training of probe in \textsc{MappedLangs} setting on top of the 7th layer.}
    \label{tab:l-6-separation}
\end{table}

\section{Probing setup}

We use the same setup for training the \orthoprobe~as \citet{limisiewicz-2020-introducing}, i.e. Adam optimizer \cite{kingma-2014-adam}, initial training rate $0.02$, and learning rate decay. We use \emph{Double Soft Orthogonality Regularization} to coerce orthogonality of the matrix $V$.

\subsection{Number of Parameters}

A \scalingvec~for each of $4$ objectives has a size $768 \times 1$ and an \orthotransf~for each language is a matrix of size $768 \times 768$. In \textsc{MappedLangs}, our largest memory-wise setting, we train $8$ \orthotransf \emph{s}. In this configuration, our probe has  $4,721,664$ parameters.

\subsection{Computation Time}

We optimized probes on a GPU core \textit{GeForce GTX 1080 Ti}. Training a probe in \textsc{MappedLangs} configuration takes about 3 hours. 

\section{Supplementary Results}

\subsection{UUAS results}

The \cref{tab:zs-fs-uuas} contains the results for undirected dependency trees. We use the same probing setting as in Section 3.2
without assigning directions to the edges. Similarly to \citet{chi-etal-2020-finding}, we exclude punctuation from the evaluation. 
\subsection{Validation Results}

In \cref{tab:speramn-dev}, we present the validation results corresponding to the test results in Table 1 of the main paper.

\begin{table}[!h]
\small
\begin{tabular}{@{}l|c|ccccc@{}}
\toprule
           & N           & ZH    & EU    & SL    & FI    & AR    \\ \hline
\textcolor{blue}{Chi et al.}  & \multirow{3}{*}{\rotatebox[origin=c]{90}{0}}    & 51.30 & -     & -     & 70.70 & 70.40 \\
\textsc{MappedL} &                       & 39.99 & 46.96 & 41.58 & 43.91 & 40.95 \\
\textsc{AllL}  &                       & 57.82 & 64.59 & 75.06 & 68.70 & 68.70 \\ \midrule
\textsc{MappedL} & \multirow{2}{*}{\rotatebox[origin=c]{90}{10}}   & 42.37 & 47.06 & 41.07 & 46.38 & 36.81 \\
\textsc{AllL}  &                       & 58.06 & 64.65 & 75.30 & 69.06 & 68.59 \\ \midrule
\textsc{MappedL} & \multirow{2}{*}{\rotatebox[origin=c]{90}{50}}  & 51.64 & 56.67 & 59.34 & 53.53 & 57.77 \\
\textsc{AllL}  &                       & 58.73 & 65.18 & 74.99 & 69.08 & 68.81 \\ \midrule
\textsc{MappedL}& \multirow{2}{*}{\rotatebox[origin=c]{90}{100}}  & 62.36 & 62.44 & 64.51 & 57.95 & 62.36 \\
\textsc{AllL}  &                       & 68.71 & 66.00 & 75.16 & 68.97 & 68.71 \\ \midrule
\textsc{MappedL} & \multirow{2}{*}{\rotatebox[origin=c]{90}{1000}} & 66.43 & 70.50 & 76.10 & 67.08 & 68.85 \\
\textsc{AllL}  &                       & 62.36 & 68.60 & 76.79 & 69.73 & 69.57 \\ \bottomrule
\end{tabular}
\caption{UUAS of extracted dependency trees in zero- and few-shot setting. The result of \emph{Structural Probe}~reported by \citet{chi-etal-2020-finding} for reference.}
\label{tab:zs-fs-uuas}
\end{table}

\begin{table*}[!ht]
\centering
\begin{tabular}{@{}l|ccccccccc@{}}
\toprule
\textbf{Approach} & \textbf{EN} & \textbf{ES} & \textbf{SL} & \textbf{ID} & \textbf{ZH} & \textbf{FI} & \textbf{AR} & \textbf{FR} & \textbf{EU} \\ \midrule
\multicolumn{10}{c}{\textbf{Dependency Distance Spearman's Correlation}} \\ \midrule
\textsc{In-Lang} & .816 & .861 & .844 & .822 & .815 & .803 & .835 & .872 & .749 \\
$\Delta$ \textsc{MappedLangs} & .000 & -.002 & .000 & .001 & -.001 & -.001 & -.002 & -.002 & .002 \\ 
$\Delta$ \textsc{AllLangs} & -.001 & -.007 & -.004 & -.011 & -.041 & .000 & -.022 & -.010 & -.021 \\ \midrule
\multicolumn{10}{c}{\textbf{Dependency Depth Spearman's Correlation}} \\ \midrule
\textsc{In-Lang} & .847 & .868 & .857 & .853 & .837 & .807 & .864 & .893 & .786 \\ 
$\Delta$ \textsc{MappedLangs} & -.003 & -.002 & -.004 & .000 & .002 & -.005 & -.002 & -.003 & -.001 \\ 
$\Delta$ \textsc{AllLangs} & -.004 & -.005 & -.004 & -.013 & -.034 & -.004 & -.027 & -.007 & -.033 \\ \midrule
\multicolumn{10}{c}{\textbf{Lexical Distance Spearman's Correlation}} \\ \midrule
\textsc{In-Lang} & .898 & .880 & .867 & .857 & .777 & .664 & .726 & .810 & .714 \\ 
$\Delta$ \textsc{MappedLangs} & .000 & .001 & -.001 & .003 & .001 & .001 & .027 & .008 & -.005 \\ 
$\Delta$ \textsc{AllLangs} & -.005 & -.005 & -.017 & -.009 & -.001 & -.053 & .004 & -.024 & -.082 \\ \midrule
\multicolumn{10}{c}{\textbf{Lexical Depth Spearman's Correlation}} \\ \midrule
\textsc{In-Lang} & .844 & .882 & .792 & .869 & .862 & .784 & .884 & .879 & .847 \\ 
$\Delta$ \textsc{MappedLangs} & .010 & .002 & .009 & -.013 & .006 & .020 & .011 & -.006 & .012 \\ 
$\Delta$ \textsc{AllLangs} & -.010 & -.067 & -.079 & -.108 & -.055 & .000 & -.259 & -.043 & -.034 \\ \midrule
\end{tabular}
\caption{Validation Spearman's correlation between gold and predicted depths and distances.  We probe the representations of 7th layer for dependency information and representations of 5th layer for lexical information.}
\label{tab:speramn-dev}
\end{table*}

\end{document}